\documentclass[10pt,twocolumn,letterpaper]{article}
\PassOptionsToPackage{table}{xcolor}

\usepackage{cite}
\usepackage{cvpr}              %
\usepackage{lineno}
\usepackage[table]{xcolor}
\usepackage{multirow}
\usepackage{float}

\definecolor{cvprblue}{rgb}{0.21,0.49,0.74}
\usepackage[pagebackref,breaklinks,colorlinks,allcolors=cvprblue]{hyperref}

\def\paperID{6757} %
\def\confName{CVPR}
\def\confYear{2025}

\title{GIFStream: 4D Gaussian-based Immersive Video with Feature Stream
}

\makeatletter
\def\blfootnote{\gdef\@thefnmark{}\@footnotetext}
\makeatother
\author{
Hao Li, 
Sicheng Li, 
Xiang Gao, 
Abudouaihati Batuer, 
Lu Yu$^{**}$,
Yiyi Liao$^{*}$
\\
Zhejiang University
}

\newcommand{\bx}{\mathbf{x}}

\newcommand{\bS}{\mathbf{S}}
\newcommand{\bs}{\mathbf{s}}

\newcommand{\bV}{\mathbf{V}}

\newcommand{\bq}{\mathbf{q}}
\newcommand{\bR}{\mathbf{R}}

\newcommand{\bff}{\mathbf{f}}

\newcommand{\bo}{\mathbf{o}}

\newcommand{\bc}{\mathbf{c}}

\newcommand{\bT}{\mathbf{T}}
\newcommand{\bQ}{\mathbf{Q}}

\newcommand{\br}{\mathbf{r}}

\newcommand{\bPhi}{\boldsymbol{\Phi}}
\newcommand{\bmu}{\boldsymbol{\mu}}

\newcommand{\btau}{\boldsymbol{\tau}}

\newcommand{\nR}{\mathbb{R}}

\newcommand{\cL}{\mathcal{L}}

\makeatletter
\DeclareRobustCommand\onedot{\futurelet\@let@token\@onedot}
\def\@onedot{\ifx\@let@token.\else.\null\fi\xspace}

\def\Fig{Fig\onedot}   
\makeatother

\newcommand{\figref}[1]{\Fig~\ref{#1}}

\renewcommand{\eqref}[1]{Eq.~\ref{#1}}
\newcommand{\tabref}[1]{Table~\ref{#1}}

\newcommand{\boldparagraph}[1]{\vspace{0.2cm}\noindent{\bf #1:} }

\newcommand{\METHOD}{GIFStream }

\begin{document}

\maketitle
\begin{abstract}
Immersive video offers a 6-Dof-free viewing experience, potentially playing a key role in future video technology. Recently, 4D Gaussian Splatting has gained attention as an effective approach for immersive video due to its high rendering efficiency and quality, though maintaining quality with manageable storage remains challenging. To address this, we introduce GIFStream, a novel 4D Gaussian representation using a canonical space and a deformation field enhanced with time-dependent feature streams. These feature streams enable complex motion modeling and allow efficient compression by leveraging temporal correspondence and motion-aware pruning. Additionally, we incorporate both temporal and spatial compression networks for end-to-end compression. Experimental results show that GIFStream delivers high-quality immersive video at 30 Mbps, with real-time rendering and fast decoding on an RTX 4090. Project page: \color{cyan}{https://xdimlab.github.io/GIFStream}
\end{abstract}
    
\blfootnote{$^*$ Corresponding author. $^{**}$ Co-corresponding author.}
\section{Introduction}

Immersive video, which enables users to explore dynamic scenes with six degrees of freedom (6-DoF), offers a highly engaging experience with applications in virtual meetings, sports viewing, and gaming. Recently, 3D Gaussian splatting (3DGS)~\cite{kerbl20233d} and its 4D extensions have emerged as promising approaches for immersive video, achieving high-quality reconstruction and real-time rendering performance. However, to make these technologies feasible for real-world applications, it is essential to balance \textit{rendering quality} with \textit{storage requirements}. Although 4D Gaussian representations show promise for immersive video, current techniques struggle to achieve this balance effectively.

\begin{figure}[t]
    \centering
    \includegraphics[width=1.05\linewidth]{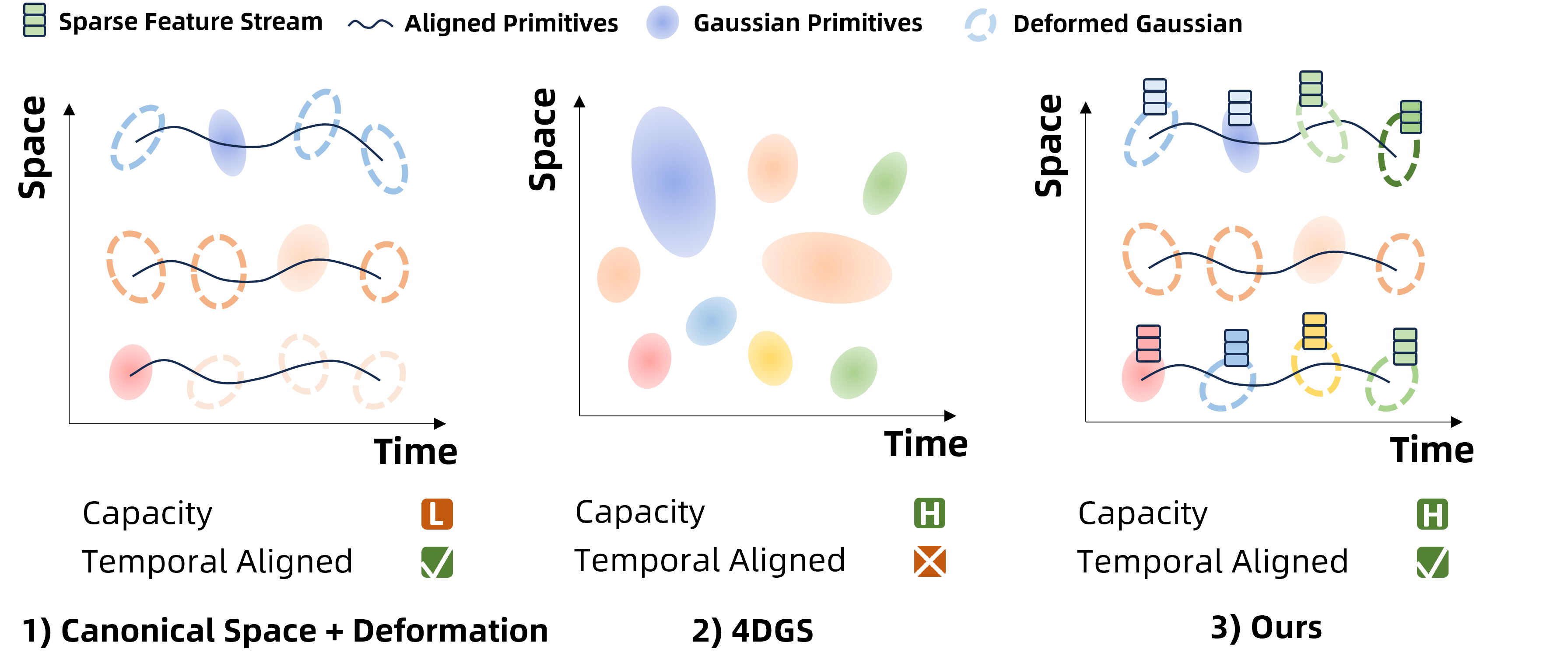}
    \caption{\textbf{Comparison of 4D Representations}. 1) Deformation-based representation stores a 3D Gaussian in a canonical space and its deformation along a long time horizon. and 2) 4D Gaussian representation which models a windowed spacetime region. We propose 3) \METHOD by adding time-dependent feature streams on top of deformation-based representation, improving its capacity while maintaining temporal alignment for efficient compression.
    }
    \label{fig:representation}
\end{figure}

To tackle this challenge, we argue that both the \textit{representation} and \textit{compression} should be considered together. An ideal 4D representation should support high-quality novel view synthesis for dynamic scenes while allowing for efficient compression. 
One line of works, as shown in \figref{fig:representation} left, represents the 4D scene as a 3D canonical space combined with a deformation field, parameterized by models such as MLPs, triplanes, per-frame single embedding or deformation bases~\cite{yang2024deformable, wu20244d, kratimenos2025dynmf, gao2024gaussianflow, huang2024sc, bae2024per}. Since the deformation field is generally lightweight, compact memory usage can be achieved by applying 3D compression methods to the canonical space. However, deformation-based methods struggle with highly dynamic scenes, as existing deformation representations lack the capacity to capture rapid motion. Another line of works~\cite{yang2023real, li2024spacetime, duan20244d}, shown in \figref{fig:representation} middle, employs 4D Gaussians to represent the scene, where each primitive models a local spatial region across a brief temporal segment. This allows for high-quality representation of complex motion using numerous 4D Gaussian primitives, though at the cost of substantial memory consumption.  Moreover, the 4D Gaussian primitives are discretely distributed across the 4D space, lacking correspondences between different primitives. This limits compression efficiency, 
since temporal redundancies could not be effectively eliminated without temporal correspondence.

In this work, we introduce \textbf{GIFStream}, a representation that captures highly dynamic content while enabling efficient temporal compression. Our key idea is to incorporate an adaptive sparse feature stream into deformation-based methods, enhancing their ability to model complex motion while enabling efficient immersive video compression through the inherent temporal correspondence as shown in \figref{fig:representation} right. 
Our canonical space consists of a set of anchor points following Scaffold-GS~\cite{lu2024scaffold}. And the key to \METHOD is that each anchor consists of a time-independent feature and a time-dependent feature stream, both are combined to predict time-dependent 3D Gaussian primitives at various timestamps, facilitating the capture of fast motion.
While adding the feature streams increases the parameter size, they are learned in a motion-adaptive manner and hence can be easily pruned for static regions.  Furthermore, it allows for efficient compression by leveraging the temporal correspondence information. 
To effectively compress the whole representation, 
we sort both time-independent and time-dependent features into two distinct video sequences, enabling the application of different video codecs including HEVC and learning-based methods. For optimal compression efficiency, we train our representation with quantization aware training and jointly train an auto-regressive entropy estimation network to conduct End-to-End compression.

To evaluate GIFStream, we conducted experiments on several challenging dynamic datasets, comparing our method with other 4D representation and compression techniques. The results indicate that \METHOD can efficiently represent highly dynamic 1080p immersive videos at a bit rate of 30 Mbps, which is comparable to that of 4K 2D videos. Furthermore, \METHOD achieves an optimal balance between quality and storage requirements, while enabling fast decoding and rendering on the RTX 4090 GPU.

\begin{figure*}[th]
    \centering
    \includegraphics[width=\linewidth]{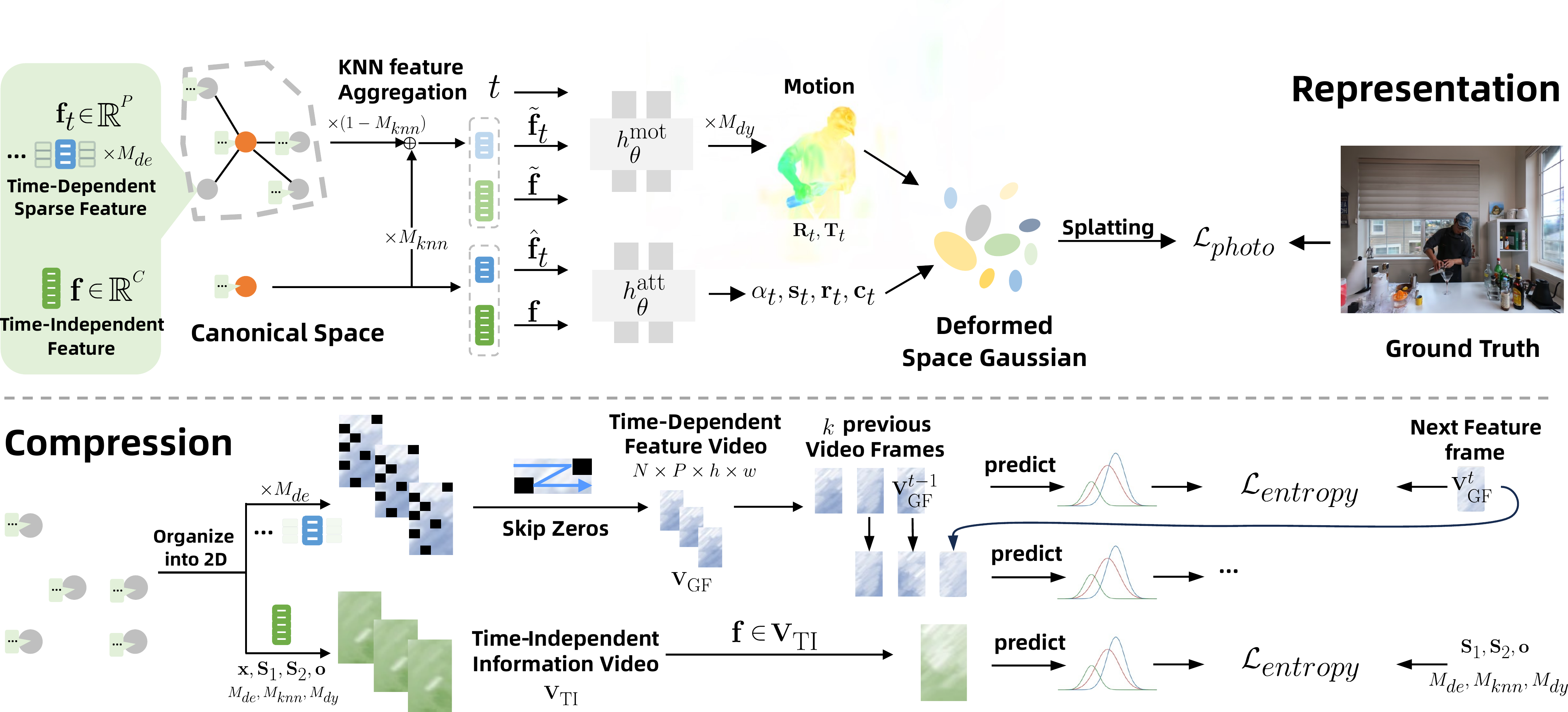}
    \caption{\textbf{Method}. (\textbf{I. Representation}) We propose enhancing deformation-based dynamic Gaussian representation using time-dependent feature streams.
    We attach time-dependent feature $\bff_t$ and time-independent feature $\bff$ to a set of anchor points. These features are aggregated to decode deformation motion $\bR_t, \bT_t$ and Gaussian attribute $\alpha_t, \bs_t, \br_t, \bc_t$ at a specific timestamp $t$ through MLPs. Finally, we render the target view through splatting. (\textbf{II. Compression}) For compression, we reorganize both time-dependent and time-independent parameters into two videos. The feature streams are first pruned and then compressed in an auto-regressive manner, effectively leveraging the temporal correspondence information. During training, we jointly optimize the rendering loss $\cL_{photo}$ and add an entropy constraint $\cL_{entropy}$. 
    }
    \label{fig:pipeline}
\end{figure*}

\section{Related Work}
\boldparagraph{NeRF-based Dynamic Novel View Synthesis} Synthesizing novel views of a dynamic scene is a challenging task. Inspired by the success of the Neural Radiance Field (NeRF)~\cite{mildenhall2020nerf}, several approaches attempt to extend it to dynamic scenes. Deformation-based works~\cite{pumarola2021d,li2022neural} construct deformation field to adjust the sampled query points for a specific time. Kplanes~\cite{fridovich2023k} and Hexplanes~\cite{cao2023hexplane} propose to directly construct 4D feature space and query the corresponding feature for specific time and space positions. Nerfplayer~\cite{song2023nerfplayer} combines the two by deforming a static field and add a time variable field like the second line. This combination improves the ability to capture new elements compared to pure deformation methods while also reducing redundancy. 
Inspired by NeRFPlayer, we propose incorporating time-dependent features to predict 3D Gaussian primitives at each timestamp, achieving higher quality while maintaining small storage requirements by further exploiting compression techniques. 

\boldparagraph{3DGS-based Dynamic Novel View Synthesis}
As 3D Gaussian Splatting achieves great quality and rendering speed, similar ideas are applied to extend 3D Gaussian Splatting to dynamic scenes. 4D Gaussian Splatting (4DGS) methods~\cite{yang2023real, li2024spacetime, duan20244d} keep Gaussian primitives in 4D space, with each primitive capturing a localized period in time and space. As illustrated in \figref{fig:representation}, these methods can be viewed as 3D Gaussian spanned over learned time periods. This yields a high capability to model dynamic scenes but at the cost of large storage. Another line of work~\cite{yang2024deformable, wu20244d, kratimenos2025dynmf, gao2024gaussianflow, huang2024sc, bae2024per}, as illustrated in \figref{fig:representation} left, basically applies deformation idea. The long-term tracking for all frames reduces the temporal redundancy but faces challenges in modeling complex dynamic content due to limitations in the deformation field's capacity. Our representation demonstrates a greater ability to capture dynamic scenes compared to existing deformation methods, and it is more compression-friendly than 4DGS due to its temporally aligned structure.

\boldparagraph{Compression for NeRF-based Methods} NeRF has emerged as a promising solution for immersive video, leveraging compression techniques to balance quality and storage efficiency effectively. Related works consistently integrate quantization, transform coding, and entropy coding methods to optimize compression performance. Early efforts focused primarily on static scenes. VQAD, Masked Wavelet NeRF, and BIRF~\cite{takikawa2022vqad, rho2023maskedwavelet, shin2023birf} introduce scalar quantization or vector quantization during training, while VQRF~\cite{li2023vqrf} applies quantization as the post-training process.
Masked Wavelet NeRF and NeRFCodec~\cite{li2024nerfcodec} utilize DCT transform or VAE-based transform networks to improve transform coding. VQRF and Masked Wavelet NeRF additionally learn a probability model to conduct entropy coding with more accurate distribution estimation. For compressing dynamic radiance fields, ReRF and videoRF~\cite{wang2023neural, wang2024videorf} introduce residual radiance fields or sequential fields, leveraging temporal relationships to enhance compression. Despite achieving a small size, these NeRF-based methods struggle to achieve real-time rendering.

\boldparagraph{Compression for 3DGS-based Methods} 
With the significant improvements in rendering speed achieved by 3D Gaussian Splatting (3DGS), compression techniques for 3DGS-based methods have quickly emerged~\cite{lee2024compact, morgenstern2023compact, navaneet2023compact3d, niedermayr2024compressed, girish2023eagles, chen2025hac, fan2023lightgaussian}. Beyond conventional quantization, transform, and entropy coding, compact 3DGS~\cite{lee2024compact} and light Gaussian~\cite{fan2023lightgaussian} reduce storage by pruning redundant Gaussian primitives. \cite{niedermayr2024compressed, morgenstern2023compact} apply Morton Sort or Self-Organizing Sort to leverage spatial relationships for compression. HAC~\cite{chen2025hac} proposes end-to-end compression using a binary hash feature grid as a context model. As for the dynamic Gaussian compression, CSTG~\cite{lee2024compact2} compresses STG using spatial compression methods. Mega~\cite{zhang2024mega} makes 4D Gaussian Splatting more compact with deformation methods. V$^3$~\cite{wang2024v} trains 3DGS in a per-frame manner using a consistent number of Gaussians, which may struggle to represent new content. While they also exploit temporal correspondence for compression, this is achieved by leveraging traditional video compression. In contrast, our proposed deformation-based representation augmented by feature streams allows for modeling new content and fast motion, and we additionally consider end-to-end temporal compression.

\section{Method}
In this section, we introduce our methods in two parts: representation(~\cref{sec:representation}) and compression(~\cref{sec:compression}), with an overview provided in \figref{fig:pipeline}. Our representation consists of a canonical space and a deformation field augmented by time-dependent feature streams.
For compression, we project both the time-dependent features and time-independent feature streams into 2D videos, enabling coding with video codecs. We train our representation with quantization-aware training combining rendering loss and entropy estimation for end-to-end compression. Additionally, 2D conventional video codecs can be applied.
More details are introduced below.

\subsection{4D Scene Representation}
\label{sec:representation}
Inspired by Scaffold-GS, we store features on a set of anchor points and decode each anchor into $K$ Gaussian primitives using small MLP. To model dynamic scenes, we propose to store both time-independent and time-dependent features at each anchor point, and we refer to the time-dependent features as \textit{feature streams}. At each timestamp, the Gaussians are decoded using two heads, one for attributes and one for motion, respectively.

\boldparagraph{Motion-Adaptive Feature Stream}
Several works explored extending 3D Gaussian Splatting to dynamic scenes using deformation methods~\cite{yang2024deformable, wu20244d, kratimenos2025dynmf, gao2024gaussianflow, huang2024sc, bae2024per}. Existing deformation-based methods primarily deform Gaussian Primitives based on positional encoding or low-dimensional feature planes. However, these methods face challenges in capturing the fine details of highly dynamic scenes, as the deformation field often lacks the capacity to handle fast motion scenarios. To solve this problem, we attach each anchor with two features: \textit{time-independent feature} \(\mathbf{f}\in \nR^{C}\) and  \textit{time-dependent feature} \(\mathbf{f}_t\in \nR^{P}\) where \(t\) indicates the time stamps. This means each anchor has a feature with \(C+N\cdot P\) channels when modeling $N$ timestamps. 
While adding time-dependent features increases the number of parameters, we consider two strategies to effectively lower storage: 1) We  make the time-dependent feature motion adaptive to eliminate them for static regions; and 2) We apply further compression by leveraging temporal correspondence, as detailed in \cref{sec:compression}.

To make the time-dependent feature motion-adaptive, we scale the feature with learnable parameters \(M_{de}\) as in \eqref{eq:anchor_mask}. 
\begin{align}
\label{eq:anchor_mask}
\hat{\bff}_{t} = M_{de}\cdot \mathbf{f}_{t}
\end{align}
We encourage \(M_{de}\) close to zero to avoid the meaningless deformation for the static anchors, and reduce the storage of feature streams. Our experiments demonstrate that the time-dependent features exhibit significant sparsity: in challenging scenes, approximately 30\% of the anchors require retention of time-dependent features, while in simpler scenes, only about 0.3\% of the anchors need to retain these features. This indicates that our representation can be effectively adapted to different scenarios.

\boldparagraph{Gaussian Attribute Prediction Head}
At each timestamp $t$, we predict the opacity, scaling, rotation, and color of $K$ Gaussian primitives from the anchor features with $h_{\theta}^\text{att}$:
\begin{equation}
    h_{\theta}^\text{att}: [\bff; \hat{\bff}_t] \mapsto \left\{\alpha^i_t, \bs^i_t, \br^i_t, \bc^i_t\right\}_{i=1}^{K} 
\end{equation}
where $h_{\theta}^\text{att}$ denotes the Gaussian attribute prediction head parameterized as small MLP. Specifically, $\alpha^i_t$ denotes the opacity, $\bs^i_t$ denotes the scaling, $\br^i_t$ denots the rotation and $\bc^i_t$ denotes the color of Gaussian $i$. Here, the $\bs^i_t$ is scaled by a per-anchor factor $\bS_1$ following Scaffold-GS.

\boldparagraph{Gaussian Motion Prediction Head}
In most scenarios, the motion of the Gaussians exhibits local smoothness. To capture this local smoothness prior, 
we aggregate the neighboring features using K-nearest neighbors (KNN), as in \figref{fig:pipeline}, before decoding them to motion vectors. To further model non-smooth motion, we introduce a learnable factor \(M_{knn}\), which enables a combination of coarse and fine motion representation:
\begin{align}
    \tilde{\mathbf{f}}_{t} &=(1-M_{knn})\sum_{k\in \mathbb{N}} \hat{\bff}_{k,t} + M_{knn}\hat{\bff}_{t} \\
    \tilde{\mathbf{f}} &=(1-M_{knn})\sum_{k\in \mathbb{N}}\bff_{k} + M_{knn}\bff
\end{align}
The obtained locally smoothed feature $\tilde{\bff}_t$ is mapped to motion combined with the time-independent feature $\bff$:
\begin{equation}
\label{eq:motion}
    h_{\theta}^\text{mot}: [\tilde{\bff}; \tilde{\bff}_t; t] \mapsto \mathbf{R}_t,\mathbf{T}_t 
\end{equation}
Here, we predict the motion \( \bR_t, \bT_t \in SE(3)\) of each anchor considering the Gaussian Primitives are tightly attached to the anchor. To ensure that anchors representing static objects remain stationary, we attach a learnable parameter \(M_{dy}\) that modulates the scaling of motion for each anchor and apply regularization on \(M_{dy}\). Specifically, the $h_{\theta}^\text{mot}$ predict two vectors $\bq_t$ and $\btau_t$, and the motion can be computed as:
\begin{align}
    \bR_t &= \mathrm{q2m}(\bar{\bq}+M_{dy}\bq_t)\\
    \bT_t &= M_{dy}\btau_t
\end{align}
where $\mathrm{q2m}$ indicates the quaternion to rotation matrix transform, $\bar{\bq}$ denotes the unit quaternion.
\begin{figure}
    \centering
    \includegraphics[width=\linewidth]{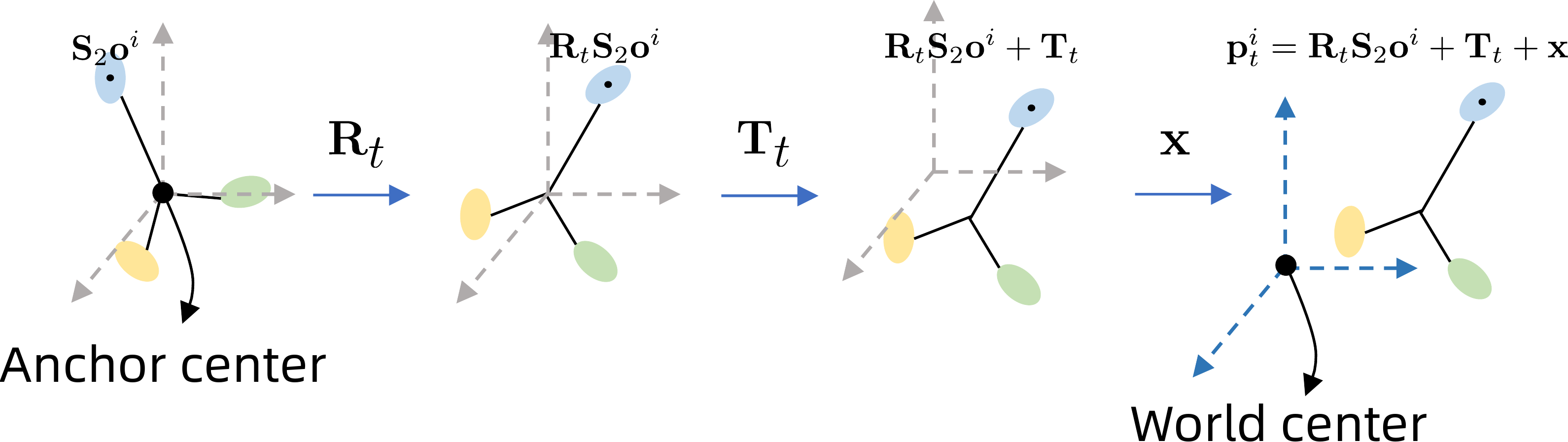}
    \caption{\textbf{Motion Illustration}. We predict the rotation and translations on the local coordinate system of the anchor.}
    \label{fig:motion_explain}
\end{figure}

Obtaining the motion of anchors, the position \(\mathbf{p}_{t}\) of generated Gaussians can be calculated as in \eqref{eq:gs_position}, where \(\mathbf{o}^i\) is the offset of $i$th Gaussian primitive relative to the anchor center, $\bS_2$ is the scaling factor of offset and \(\bx\) is the position of anchor in global coordinate. We illustrate the motion of anchor in \figref{fig:motion_explain}.
\begin{equation}
    \label{eq:gs_position}
    \mathbf{p}_{t}^i = \mathbf{R}_{t}{\bS_2 \bo^i} + \bx + \mathbf{T}_{t}
\end{equation}
Here, ${\bo^i}$, $\bS_2$ and $\bx$ are additionally time-independent parameters to be transmitted.

\subsection{Compression}
\label{sec:compression}
\boldparagraph{Sorting-based 2D re-organization}
Leveraging the alignment provided by the canonical space and deformation, our representation can be sorted and reorganized into two videos as shown in \figref{fig:pipeline}, allowing both temporal and spatial compression. Specifically, we map the Gaussian primitives from 3D to 2D space following~\cite{morgenstern2023compact}. While \cite{morgenstern2023compact} is applied to static scene compression and apply sorting based on position and color, our sorting is conducted according to the position of the canonical anchors and the PCA of the time-independent features. More details can be seen in our supplementary material. 

Obtaining the 2D position of each anchor, we stack the parameters of each anchor into two videos. One video \(\bV_{\text{TI}} \in \nR^{ (12+3\times K+C) \times 1\times H \times W} \) contains time-independent information $\{\bx\in\nR^3,\bS_1\in\nR^3,\bS_2\in\nR^3,\{\bo_i\in\nR^3\}_{i=1}^{K},M\in\nR^3,\bff\in\nR^C\}$, where $(12+3\times K+C)$ is the number of channels and is considered as the time dimension, and $M=\{M_{de}, M_{knn}, M_{dy}\}$. The other video \(\bV_{\text{GF}}\in \nR^{N \times P \times h \times w } \) contains time-dependent feature stream $\{\bff_t\}_{t=1}^{N}$, where the video consists of $N$ frames, each with $P$ channels.

Observing the sparsity of the feature stream, we directly discard the feature stream set to zero by \(M_{de}\), and reorganize remaining features by skipping over the zero using line-by-line scanning. As a result, the resolution of \(\bV_{\text{GF}}\) becomes much smaller than that of \(\bV_{\text{TI}}\), i.e., $h<H$ and $w<W$. Note that we transmit  \(M_{de}\) to preserve the point-wise correspondence between \(\bV_{\text{GF}}\) and \(\bV_{\text{TI}}\). Following reorganization, we design an end-to-end compression method containing quantization-aware training, entropy constraint and entropy encoding. Notably, the representation trained with our entropy supervision can also be compressed using traditional video codecs, such as VVC~\cite{bross2021vvcoverview} and HEVC~\cite{sullivan2012hevcoverview}. 

\boldparagraph{Quantization-Aware Training}
To simulate compression process, we quantize the features into integer values during training, utilizing the Straight-Through Estimator (STE): 
\begin{align}
\label{eq:ste}
\bar{\mathbf{f}} = \mathrm{SG}\left(\mathrm{round}(\mathbf{f}) - \mathbf{f}\right) + \mathbf{f}
\end{align}
where \(\mathrm{SG()}\) denotes gradient-stopping operation. This approach ensures effective feature quantization while maintaining the differentiability needed for end-to-end training.

\boldparagraph{Entropy Regularization}
We use conditional entropy regularization to compress both $\bV_{\text{TI}}$ and $\bV_{\text{GF}}$ inspired by End-to-End compression methods~\cite{lu2019dvc, li2021deep, mentzer2022vct, balle2018variational, chen2025hac}. 
We compress the time-dependent feature stream video $\bV_{\text{GF}}$ and the time-independent video $\bV_{\text{TI}}$ separately.

For entropy regularization of $\bV_{\text{GF}}$, we exploit auto-regressive probability estimation inspired by video compression methods~\cite{lu2019dvc, li2021deep, mentzer2022vct}. Let $\bV^t_{\text{GF}}\in \nR^{P\times h \times w}$ denote the $t$th frame of $\bV_{\text{GF}}$. To estimate the entropy for every frame, we jointly train neural networks to predict the distribution of the next frame $\bV^t_{\text{GF}}$ with \(k\) previous frames $\{\bV^{t-k}_{\text{GF}},\cdots,\bV^{t-1}_{\text{GF}}\}$ as input:
\begin{align}
\label{eq:dis_predict}
h_{\theta}^\text{ent}: [\bV^{t-k}_{\text{GF}}; \cdots; \bV^{t-1}_{\text{GF}}] \mapsto \{\bmu_t, \boldsymbol{\sigma}_t\}
\end{align}
where $h_{\theta}^\text{ent}$ is convolutional neural networks, $\bmu_t$ and \(\boldsymbol{\sigma}_{i}\) are the predicted mean and variance of frame $\bV^{t}_{\text{GF}}$ assuming Gaussian distribution. If available previous frames are fewer than \(k\), we directly pad the input with zeros.
Based on this predicted distribution, we compute the prior probability of the next frame and subsequently estimate the entropy loss \(\cL_{entropy}\). 
Given the predicted \(\sigma_t\) and $\bmu_t$, the entropy loss of frame $\bV^{t}_{\text{GF}}$ is estimated as shown in \eqref{eq:entropy}, where $\bPhi_{\boldsymbol{\sigma}_t}$ denotes the cumulative distribution function.
\begin{align}
\label{eq:entropy}
\cL_{entropy} = -\sum_t\text{log}(&\bPhi_{\bmu_t,\boldsymbol{\sigma}_t}(\bV^{t}_{\text{GF}}+0.5) - \\ &\bPhi_{\bmu_t,\boldsymbol{\sigma}_t}(\bV^{t}_{\text{GF}}-0.5))\notag
\end{align}
where $0.5$ indicates half of the quantization step as we quantize features into integer values.

The entropy regularization for $\bV_{\text{TI}}$ follows a similar approach to $\bV_{\text{GF}}$. For $\bff$ in $\bV_{\text{GF}}$, we group feature frames into fragments and perform probability estimation conditioned on the $k$ previous fragments. The probability for the remaining attributes $\{\bx,\bS_1,\bS_2,\{\bo_i\}_{i=1}^{K},M\}$ in $\bV_{\text{TI}}$ is estimated based on $\bff$, as we empirically observe that $\bff$ correlates with these attributes. More details can be seen in the supplementary material. 

\boldparagraph{Encoding Methods}
After quantization-aware training with entropy constraints, we conduct quantization and entropy coding using the learned probability distribution, typically known as end-to-end compression.
Specifically, we quantize parameters with the quantization step used in training and conduct entropy encoding using rANS~\cite{duda2013asymmetric} codec. 

Optionally, we can also utilize the existing video codec to conduct compression. Although the compression efficiency is lower than the end-to-end compression, their advantage lies in their widespread support and specialized hardware acceleration. Note that the quantization-aware training and entropy regularization during training also contribute to improved compression performance when using these traditional video codecs, as the distribution of the representation is constrained.

\subsection{Optimization Strategy}
\boldparagraph{Densification and Pruning}
The densification of the vanilla Scaffold-GS~\cite{lu2024scaffold} is directed by the gradients of parameters, which are averaged over several iterations during training. This averaging approach works well in 3D scenarios because it effectively reflects errors from different views. However, in 4D settings, this makes the gradients averaged across time, which results in Gaussians representing fast-moving objects receiving insufficient gradients to reach the clone or split threshold. Since these Gaussians are crucial for capturing dynamic details, we modify the gradient accumulation method. Specifically, we continue to average the gradients across different views but store the gradients at different time stamps separately. For each anchor, we compute the maximum gradient value across time and then combine it with the temporal average gradient using a weight factor \(\alpha\). As a result, the accumulated gradients are adjusted as \eqref{eq:gradients}, where $\mathbf{g}_{t}$ is the norm of gradients.
\begin{equation}
\label{eq:gradients}
\bar{\mathbf{g}}=\alpha\max_{t\in [0,1]}(\mathbf{g}_{t}) + (1-\alpha)\frac{1}{L}\sum_{t\in [0,1]}\mathbf{g}_{t}
\end{equation}

For pruning,
we follow CSTG to multiply each anchor's scaling and opacity with learnable binary mask \(M_p\), and encourage the average value of \(M_p\) to be small through regularization loss $\cL_m$. If the Gaussian primitives are deemed significant, the rendering loss generates opposing gradients to preserve these primitives. Every 500 iterations, we evaluate the mask values and prune any primitives whose corresponding mask values approach zero.

\boldparagraph{Regularization and Loss Function}
We train our representation with photo loss \(\cL_{photo}\) and regularization terms. To encourage the rigidity of Gaussian primitives and encourage the sparsity of our motion-aware feature streams, we punish the parameters \(M = \{M_{de}, M_{dy}, M_{knn}, M_{p}\}\) with regularization loss:
\begin{equation}
    \label{eq:mask_loss}
    \cL_m = |M|
\end{equation}

To enhance the temporal consistency, we attach L1 loss to the deformed attributes \(\mathbf{A}_t\) of Gaussian primitives for adjacency timestamps:
\begin{equation}
    \label{eq:smooth}
    \cL_s = |\mathbf{A}_t - \mathbf{A}_{t+1}|
\end{equation}

Following \cite{morgenstern2023compact}, we enhance the spatial smoothness of $\bV_\text{TI}$ through MSE loss $\cL_{ss}$ between the re-organized frames and their corresponding blurred versions.
In summary, we combine all the aforementioned losses with hyper-parameters $\lambda_e, \lambda_r$ as follows:
\begin{equation}
    \label{eq:loss}
    \cL=\cL_{photo}+\lambda_e \cL_{entropy} + \lambda_r (\cL_s+\cL_{ss}+\cL_m)
\end{equation}

\begin{figure}[t]
    \centering
    \begin{subfigure}{0.48\linewidth}
        \centering
        \parbox[][0.01cm][c]{1\linewidth}{\subsection*{~}\label{fig:RDPSNR}}
        \includegraphics[width=1\linewidth]{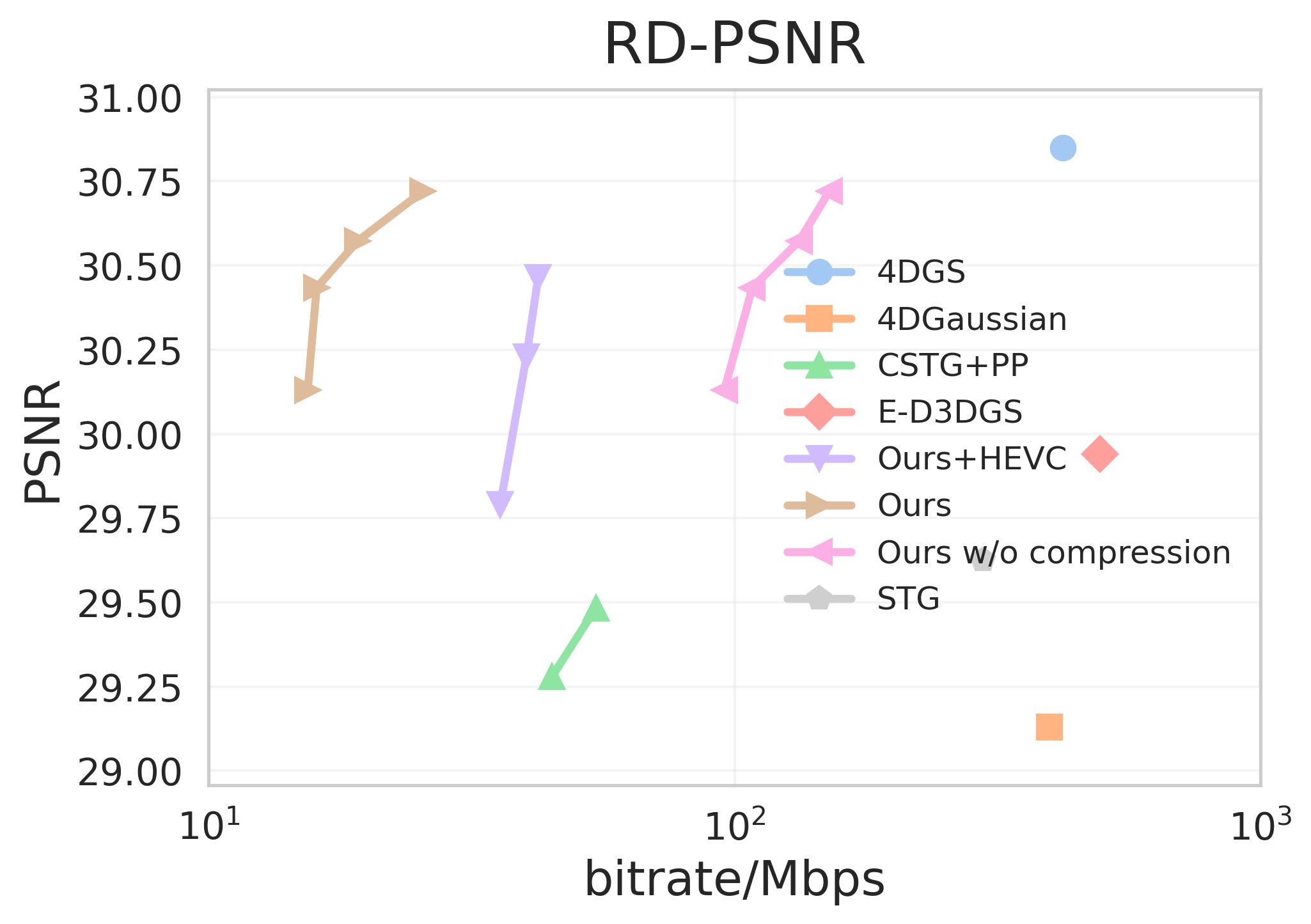}
    \end{subfigure}
    \begin{subfigure}{0.48\linewidth}
        \centering
        \parbox[][0.01cm][c]{1\linewidth}{\subsection*{~}\label{fig:RDLPIPS}}
        \includegraphics[width=1\linewidth]{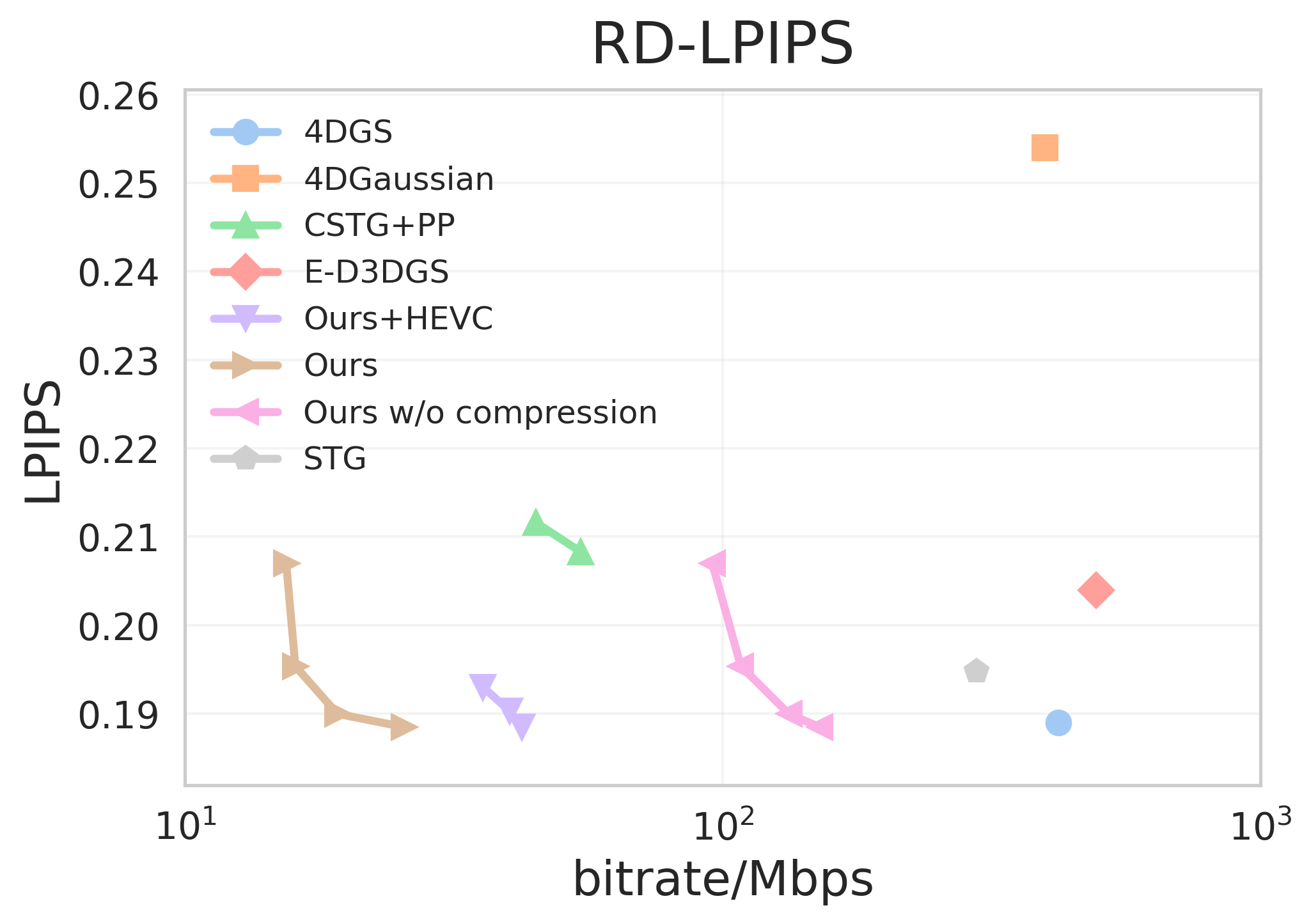}
    \end{subfigure}
    \caption{\textbf{RD Curve Comparison on MPEG dataset.} We visualize the RD Curve results in the \textbf{GOP 65} setting.}
    \label{fig:RD_Curve}
\end{figure}

\definecolor{tabfirst}{rgb}{1, 0.7, 0.7} %
\definecolor{tabsecond}{rgb}{1, 0.85, 0.7} %
\definecolor{tabthird}{rgb}{1, 1, 0.7} %

\begin{table*}[t]
\centering
\resizebox{1.0\linewidth}{!}{
\begin{tabular}{lccccccccccccccc}
\toprule

& \multicolumn{5}{c}{Neur3D}  & \multicolumn{5}{c}{Panoptic Sport$^*$} & \multicolumn{5}{c}{MPEG} \\\cmidrule(lr){2-6} \cmidrule(lr){7-11} \cmidrule(lr){12-16}

Method       & PSNR $\uparrow$ & SSIM $\uparrow$ & LPIPS (ALEX) $\downarrow$ & FPS $\uparrow$ & Storage (MB) $\downarrow$ & PSNR $\uparrow$ & SSIM $\uparrow$ & LPIPS (VGG) $\downarrow$ & FPS $\uparrow$ & Storage (MB) $\downarrow$ & PSNR $\uparrow$ & SSIM $\uparrow$ & LPIPS (VGG) $\downarrow$ & FPS $\uparrow$ & Storage (MB) $\downarrow$ \\\midrule

NeRFPlayer~\cite{song2023nerfplayer} & 30.69 & 0.932 & 0.111 & 0.05 & 5100
& - & - & - & - & -
& - & - & - & - & - \\

K-Planes~\cite{fridovich2023kplanes} & 31.63 & - & - & 0.30 & 311 
& - & - & - & - & -
& - & - & - & - & - \\

Dynamic3DGS~\cite{luiten2024dynamic} & - & - & - & - & -
& \cellcolor{tabsecond}28.84 & 0.910 & 0.175 & - & 2083.8 
& - & - & - & - & -\\

4DGaussian~\cite{wu20244d} & 31.15 & - & \cellcolor{tabthird}0.049 & 30 & \cellcolor{tabthird}90
& 27.71 & \cellcolor{tabsecond}0.914 & 0.175 & 55 & \cellcolor{tabthird}137.7
& 29.13 & 0.866 & 0.254 & 21 & 108\\

4DGS~\cite{yang2023real} & \cellcolor{tabsecond}32.01 & - & - & \cellcolor{tabthird}114 & 202 
& \cellcolor{tabthird}28.68 & \cellcolor{tabthird}0.911 & \cellcolor{tabsecond}0.157 & \cellcolor{tabthird}200 & 973.8 
& \cellcolor{tabsecond}30.50 & \cellcolor{tabsecond}0.888 & \cellcolor{tabsecond}0.191 &  \cellcolor{tabthird}80 & 114 \\

E-D3DGS~\cite{bae2024per} & 31.42 & 0.945 & \cellcolor{tabfirst}0.037 & - & 137 
& 25.61 & 0.896 & \cellcolor{tabthird}0.172 & 50 & 297.9
& \cellcolor{tabthird} 29.94 & 0.881 & 0.204 & 35 & 134 \\

STG~\cite{li2024spacetime} & \cellcolor{tabfirst}32.05 & \cellcolor{tabfirst}0.946 & \cellcolor{tabsecond}0.044 & \cellcolor{tabsecond}140 & 200 
& 25.09 & 0.900 & 0.181 & \cellcolor{tabsecond}270 & 180.9
& 29.62 & \cellcolor{tabthird}0.886 & \cellcolor{tabthird}0.195 & \cellcolor{tabsecond}93 &  \cellcolor{tabthird}80 \\

CSTG + PP~\cite{lee2024compact2} & 31.69 & \cellcolor{tabsecond}0.945 & 0.054 & \cellcolor{tabfirst}186 & \cellcolor{tabsecond}15 
& 26.13 & 0.902 & 0.192 & \cellcolor{tabfirst}360 & \cellcolor{tabsecond}23.4
&                      29.48 &                      0.885 &                      0.208 &  \cellcolor{tabfirst}115 & \cellcolor{tabsecond}15 \\

Ours       &  \cellcolor{tabthird}31.75 &  \cellcolor{tabthird}0.938 &  0.051 &                      95 &  \cellcolor{tabfirst}10 
& \cellcolor{tabfirst}29.50 & \cellcolor{tabfirst}0.931 & \cellcolor{tabfirst}0.114 & 100 & \cellcolor{tabfirst}12.6
& \cellcolor{tabfirst}30.72 & \cellcolor{tabfirst}0.892 & \cellcolor{tabfirst}0.188 & 70 & \cellcolor{tabfirst}7 \\

\bottomrule
\end{tabular}}
\caption{\textbf{Quantitative comparisons results.} Panoptic Sport$^*$ denotes that experiments are conducted only on two specific scenes: basketball and boxes. CSTG + PP refers to CSTG with post-processing techniques, including quantization, entropy encoding, and pruning.}
\label{tab:full_evaluation}
\end{table*}

\section{Experiments}

\begin{figure*}
    \centering
    \includegraphics[width=\linewidth]{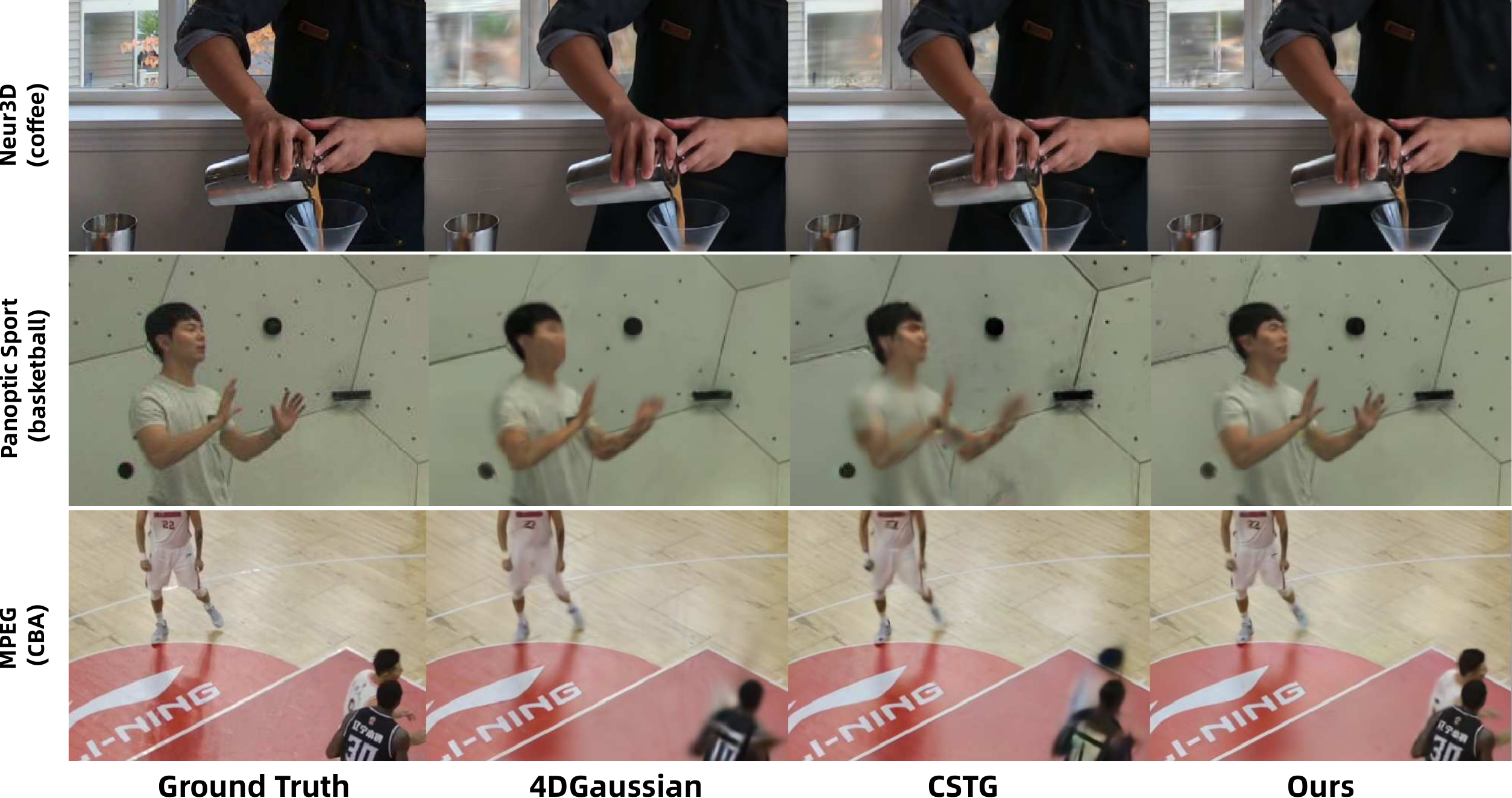}
    \caption{\textbf{Qualitative Comparison}. We implement the 4DGaussian, CSTG and compare their reconstructed quality with ours on these scenes. Our methods achieve better quality in scenes with fast motion, see the basketball player at the bottom right corner of the 3th row. }
    \label{fig:subjective}
\end{figure*}

\subsection{Experimental Settings}
\label{sec:exp_setup}
\boldparagraph{Baselines} We compare our methods with the latest 4D Gaussian compression method CSTG~\cite{lee2024compact2} and representative methods STG~\cite{li2024spacetime}, 4DGS~\cite{yang2023real} E-D3DGS~\cite{bae2024per}, and 4DGaussian~\cite{wu20244d}. Among these, STG and 4DGS represent 4D Gaussian Splatting methods, while 4DGaussian and E-D3DGS represents deformation-based methods.

\boldparagraph{Datasets} We conduct experiments on three datasets: the widely used Neur3D~\cite{li2022neural}, Panoptic Sports~\cite{joo2015panoptic} and a more challenging MPEG dataset, which is available for standardization activities related to 3DGS compression in MPEG.
The Neur3D dataset comprises six indoor multi-view sequences captured by 18-21 cameras at 2704×2028 resolution. For this dataset, we follow previous methods by using half resolution for the first 300 frames.
The Panoptic Sports dataset contains six sequences at 640×360 resolution, each captured from 31 cameras. Our experiments focus on two sequences: basketball and boxes.
The MPEG dataset includes two indoor multi-view sequences featuring fast motion scenarios (bartending and basketball), captured by 20-30 cameras at approximately 1080p resolution. We conduct experiments on the first 65 frames at original resolution.

\boldparagraph{Metrics} We compute Rate-Distortion (RD) metrics to comprehensively evaluate different methods. For distortion evaluation, we use the PSNR, SSIM, and LPIPS~\cite{ZhangIESW18lpips} as the quality metrics. For representation size, we use the bitrate, consistent with standard video compression methods.

\subsection{Implementation Details}
\label{sec:implement_details}
\boldparagraph{Initialization} We train a Group-of-Pictures (GOP) jointly, and use the sparse point clouds as the initialization generated by COLMAP from the first frame.

\boldparagraph{Hyper-parameter and Training Process} In all experiments we set \(\lambda_r\) to 0.0005 and $K$ to 5. The feature stream channel \(P\) is set to 4 for the Neur3D dataset and 8 for the MPEG dataset since the latter changes faster across time. The time-independent feature channel $C$ is set to 48 for the ``coffee martini'', ``cook spinach'', and ``flame salmon'' scenes in the Neur3D dataset for better quality, and 24 for the remaining scenes. We train every GOP with 30K iterations and first only train canonical space for 5\% iterations, and then we train the canonical space and deformation field jointly without any compression for 15\% iterations. For the remaining iterations, we reorganize Gaussian primitives every 500 iterations and introduce the 2D smoothness constraint. Additionally, we start quantization-aware training and add entropy constraints. We adjust $\lambda_e$ for the entropy loss from 0.012 to 0.00025 to obtain different bitrates for the same scene.

\subsection{Comparison with State-Of-The-Art}
\label{sec:main_results}
\boldparagraph{Quantitative Evaluation} The results demonstrate that our method achieves an optimal balance between quality and storage efficiency. In terms of quality, our approach surpasses existing deformation-based methods and achieves comparable results to 4DGS methods, which model windowed spacetime regions. Furthermore, our method delivers superior quality on both the MPEG and Panoptic Sports datasets. In terms of storage requirements, our approach achieves the smallest footprint across all evaluated datasets.
Although our rendering frame rate is lower than 4DGS methods, which do not require deformation calculations, our approach still delivers performance exceeding 60 FPS on a consumer-grade NVIDIA RTX 4090 graphics card.

\boldparagraph{Qualitative Evaluation} 
We show novel view synthesis results in \figref{fig:subjective}. The first two rows demonstrate that our method is on par with existing methods on the Neur3D dataset at a small storage cost. For the MPEG datasets, 4DGaussian and CSTG struggle to model the fast motion, leading to artifacts or blurriness. In contrast, our method is capable of providing good visual quality thanks to the time-dependent features. Please refer to our supplementary material for more comparisons.

\boldparagraph{RD-Curve Comparison} To fully evaluate compression performance, we provide 4 different rate points on the datasets and visualize the RD-Curve on \figref{fig:RD_Curve}. Lower bit rates can be achieved by adjusting currently fixed hyperparameters, such as the feature quantization step. The results further prove that our method achieves the optimal quality and storage balance of the rate-distortion performance.

\boldparagraph{Decoding Speed}
Our decoding process includes two parts: distribution prediction and entropy decoding. Since our distribution prediction MLP $h_{\theta}^\text{ent}$ is tiny and the feature stream is sparse, enabling us to complete the decoding process within one second for a two-second, 30 FPS Bartender video. The feature distribution prediction can reach 100 FPS on the NVIDIA RTX4090 GPU for the Bartender scene. We use rANS to conduct entropy decoding and it reaches 5 FPS for time-independent parameters and 200 FPS for feature streams. Since we represent 65 timestamps with a single model, the overall decoding time can be acceptable for real-time playback.

\subsection{Ablation Study}
\label{sec:ab_study}
\begin{table}[t]
\centering
\resizebox{1.0\linewidth}{!}{
\begin{tabular}{lcccc}
\toprule
\cmidrule(lr){2-5}
Method       & PSNR $\uparrow$ & SSIM $\uparrow$ & LPIPS $\downarrow$ & Storage (MB) $\downarrow$ \\\midrule

Full                  & 31.94 & 0.879 & 0.190 & 5.3 \\
per-frame Scaffold-GS & 31.96 & 0.881 & 0.184 & 1283 \\
w/o compression       & 32.13 & 0.884 & 0.184 & 46.1 \\
w/o feature stream $\mathbf{f}_t$    & 30.59 & 0.867 & 0.208 & 4.4 \\
w/o sparse mask $M_{de}$      & 31.93 & 0.879 & 0.190 & 6.5 \\
w/o KNN for $\tilde{\mathbf{f}}, \tilde{\mathbf{f}_t}$             & 31.86 & 0.877 & 0.192 & 5.7 \\
w/o our densification & 31.82 & 0.877 & 0.194 & 4.5 \\
\bottomrule

\end{tabular}}
\caption{\textbf{Ablation Study}. We evaluate different components in our methods on Bartender Scene.}
\label{tab:ab}
\end{table}

We conduct ablation studies on the Bartender scene to evaluate the contribution of each component, with results shown in \tabref{tab:ab}. We remove each part individually based on the full model. We provide a detailed memory breakdown in the supplementary material.

\boldparagraph{Compare to Scaffold-GS} We run Scaffold-GS without modifications on each frame to evaluate the representation quality and compactness of our approach (per-frame Scaffold-GS). Our observations indicate that \METHOD achieves comparable quality to this baseline while requiring only 1/200 of the storage. Furthermore, \METHOD effectively eliminates the flickering artifacts observed in per-frame Scaffold-GS.

\boldparagraph{Compression} We show the result without quantization-aware training, entropy regularization, and entropy coding (w/o compression). Here, we keep the pruning strategy of the feature streams by discarding those set to 0 by $M_{de}$, and prune the anchors with $M_{p}$. We discard around 83.5\% feature streams and the feature stream occupies 60\% storage size in the current model. We observe that the uncompressed size is 9 times larger than compressed.

\boldparagraph{Feature Stream} We evaluate the effectiveness of the feature streams by using only the time-independent feature and positional encoding of time as the input of MLPs (w/o feature stream $\bff_t$). 
The results show a clear degradation in rendering performance, highlighting the importance of time-dependent features for capturing fast motion. Meanwhile, storage remains largely unaffected, as the feature stream is sparse and further compressed into a small size.

\boldparagraph{Motion Mask} Next, we remove the mask $M_{de}$ and maintain all the feature streams (w/o sparse mask $M_{de}$). In this experiment, the bit rate of the feature stream increases around 80\% (from 1.5 MB to 2.7 MB), indicating the importance of sparse mask $M_{de}$.

\begin{figure}[t]
    \centering
    \includegraphics[width=\linewidth]{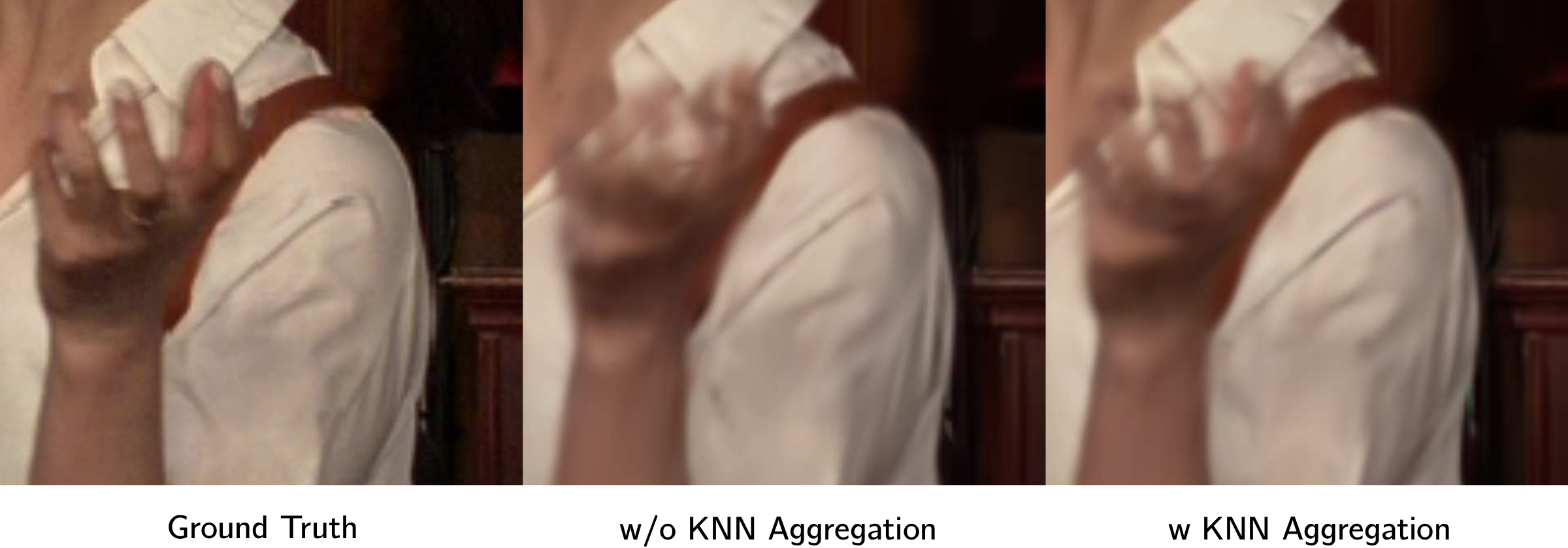}
    \caption{\textbf{Ablation Study about KNN Aggregation.} We compare the visual quality of dynamic details in this figure.}
    \label{fig:distortion}
\end{figure}

\boldparagraph{KNN Aggregation} We further investigate the KNN feature aggregation by using $\hat{\bff}, \hat{\bff}_t$ for the motion prediction head $h_{\theta}^\text{mot}$ (w/o KNN for $\tilde{\bff}, \tilde{\bff}_t$). While this modification results in only a minor drop in rendering performance in quantitative metrics, it leads to noticeable visual distortion as shown in \figref{fig:distortion}. Additionally, we observe an increase in storage requirements, as the number of primitives grows without the regularization provided by KNN.

\boldparagraph{Densification} Finally, we apply the densification strategy of Scaffold-GS and demonstrate that the rendering performance drops without using our densification strategy (w/o our densification). This is indicated by smaller storage requirements and fewer Gaussians being inserted when averaging the gradients across the time dimension.

\section{Conclusion}
We propose to incorporate a sparse feature stream to deformation-based 4D Gaussian representation, enabling capturing highly dynamic content and efficient temporal compression for time-dependent content. We also design end-to-end compression methods for our representation making the 3DGS-based immersive video's bit rate comparable to 4K 2D videos. Experiments show that our representation achieves the best rate and distortion performance compared to baseline methods, and our decoding speed is acceptable for real-time immersive video watching experience. This evidence demonstrates that our method provides a promising solution for future immersive video.

\newpage
\section*{Acknowledgements}
This work is supported by the National Natural Science Foundation of China under Grant No. 62441223, No. 62202418 and No. U21B2004. Yiyi Liao and Lu Yu are with Zhejiang Provincial Key Laboratory of Information Processing, Communication and Networking (IPCAN), Hangzhou 310007, China.
{
    \small
    \bibliographystyle{ieeenat_fullname}
    
    \bibliography{bibliography, bibliography_custom, bibliography_long, main}
}

\clearpage
\setcounter{page}{1}
\maketitlesupplementary

\setcounter{section}{0}
\section{Overview}
\label{sec:Overview}

In this supplementary material, we present three sections: (I) additional details on the methods, (II) supplementary information on the experiments, and (III) a discussion of limitations and future work.

\section{Supplementary for Methods}
\boldparagraph{Details on 3D-to-2D Sorting} We adopt the sorting strategy introduced in PLAS~\cite{morgenstern2023compact} to reorganize the Gaussian primitives from 3D space to 2D space based on the similarity of their attributes. While \cite{morgenstern2023compact} sorts static 3D Gaussian primitives using their positions, zero-degree spherical harmonic (SH) coefficients, and scaling, we sort our canonical anchors based on their positions and the three principal components of the time-independent features $\bff$ obtained through Principal Component Analysis (PCA). All the parameters used for sorting are normalized to the range $[0,1]$ along the channel dimension through min-max normalization.

\boldparagraph{Compression for Time-Independent Video}
Our time-independent video $\bV_{\text{TI}}$, comprises two components: time-independent features frames and other attributes frames, including $\bx,\bS_1,\bS_2,\{\bo^i\},M$. For compressing this video, there are many options, including existing 3DGS compression methods~\cite{lee2024compact, morgenstern2023compact, navaneet2023compact3d, niedermayr2024compressed, girish2023eagles, chen2025hac, fan2023lightgaussian} and our feature stream compression methods. We perform compression following the same principle as compressing the time-dependent video to keep our overall method simple and effectively exploit intra-channel and spatial dependencies.

\begin{table}[t]
\centering
\small
\resizebox{\linewidth}{!}{
\begin{tabular}{lcccccc}
\toprule
Methods & Rate & Scenes  & PSNR $\uparrow$ & SSIM $\uparrow$ & LPIPS (VGG) $\downarrow$ & Storage (MB) $\downarrow$ \\\midrule
\multirow{3}{*}{4DGS}   &\multirow{3}{*}{rate0}       & Bartender      & 31.58 & 0.865 & 0.221 & 126.2  \\
                        &                             & CBA            & 29.43 & 0.911 & 0.161 & 101.6  \\
                        &                             & \textbf{AVG}            & 30.50 & 0.888 & 0.191 & 113.9 \\
\midrule
\multirow{3}{*}{4DGaussian}   &\multirow{3}{*}{rate0}       & Bartender      & 31.06 & 0.858 & 0.249 & 108.0   \\
                        &                             & CBA            & 27.2 & 0.875 & 0.259 & 107.0  \\
                        &                             & \textbf{AVG}            & 29.13 & 0.867 & 0.254 & 107.5 \\
\midrule
\multirow{3}{*}{STG}   &\multirow{3}{*}{rate0}       & Bartender      & 31.4 & 0.875 & 0.207 & 49.1   \\
                        &                             & CBA            & 27.85 & 0.896 & 0.183 & 111.3  \\
                        &                             & \textbf{AVG}    & 29.63 & 0.886 & 0.195 & 80.2  \\
\midrule
\multirow{6}{*}{CSTG}   &\multirow{3}{*}{rate0}       & Bartender      & 31.12 & 0.876 & 0.218 & 10.7   \\
                        &                             & CBA            & 27.85 & 0.895 & 0.199 & 18.8  \\
                        &                             & \textbf{AVG}            & 29.48 & 0.885 & 0.208 & 14.7  \\
\cmidrule(lr){2-7}
                        &\multirow{3}{*}{rate1}       & Bartender      & 31.06 & 0.874 & 0.221 & 8.2   \\
                        &                             & CBA            & 27.5 & 0.892 & 0.202 & 16.2  \\
                        &                             & \textbf{AVG}            & 29.28 & 0.883 & 0.212 & 12.2  \\
\midrule
\multirow{12}{*}{Ours}   &\multirow{3}{*}{rate0}       & Bartender      & 31.94 & 0.879 & 0.190 & 5.3   \\
                        &                             & CBA            & 29.5 & 0.906 & 0.187 & 8.5  \\
                        &                             & \textbf{AVG}            & 30.72 & 0.893 & 0.189 & 6.9  \\
\cmidrule(lr){2-7}
                        &\multirow{3}{*}{rate1}       & Bartender      & 31.69 & 0.876 & 0.195 & 3.3   \\
                        &                             & CBA            & 29.46 & 0.906 & 0.185 & 7.1  \\
                        &                             & \textbf{AVG}            & 30.57 & 0.891 & 0.190 & 5.2  \\
\cmidrule(lr){2-7}
                        &\multirow{3}{*}{rate2}       & Bartender      & 31.48 & 0.873 & 0.201 & 2.6   \\
                        &                             & CBA            & 29.39 & 0.905 & 0.190 & 6.1  \\
                        &                             & \textbf{AVG}            & 30.43 & 0.889 & 0.195 & 4.3  \\
\cmidrule(lr){2-7}
                        &\multirow{3}{*}{rate3}       & Bartender      & 31.35 & 0.872 & 0.207 & 2.3   \\
                        &                             & CBA            & 28.91 & 0.897 & 0.207 & 6.1  \\
                        &                             & \textbf{AVG}            & 30.13 & 0.885 & 0.207 & 4.2  \\
\bottomrule

\end{tabular}}
\caption{\textbf{Per-Scene Results on The MPEG dataset}. We present the PSNR, SSIM, LPIPS (VGG), and storage result of each methods. }
\label{tab:perscene_mpeg}
\end{table}

\begin{table*}
\centering
\small
\resizebox{1\linewidth}{!}{
\begin{tabular}{lccccc}
\toprule
Experiment &  Scenes  & Time-independent Feature $\mathbf{f}$ (MB)& Attributes (MB) & Time-dependent Feature $\mathbf{f}_t$ (MB)& Neural networks  (MB)\\\midrule
\multirow{3}{*}{GIFStream w/ Compression}   & Bartender      & 0.79 & 2.85 & 1.46 & 0.1  \\
                        & CBA            & 1.47 & 5.62 & 2.88 & 0.1  \\
                        & \textbf{AVG}   & 1.13 & 4.23 & 2.17 & 0.1 \\
\midrule
GIFStream w/o Compression & Bartender      & \multicolumn{2}{c}{18.68} & 27.42 & 0.1  \\
\midrule
GIFStream w/o Sparse Mask $M_{de}$ & Bartender      & 0.88 & 2.89 & 2.68 & 0.1  \\
\bottomrule

\end{tabular}}
\caption{\textbf{Memory Breakdown}. For a more comprehensive evaluation, we present the memory breakdown for each experiment. In the GIFStream w/o Compression experiment, we provide the total memory usage of the time-independent features and attributes.}
\label{tab:breakdown}
\end{table*}

\begin{figure*}[t]
    \centering
    \hspace{-0.5cm}
    \begin{subfigure}{0.33\linewidth}
        \centering
        \parbox[][0.01cm][c]{1\linewidth}{\subsection*{~}\label{fig:RDPSNR2}}
        \includegraphics[width=1\linewidth]{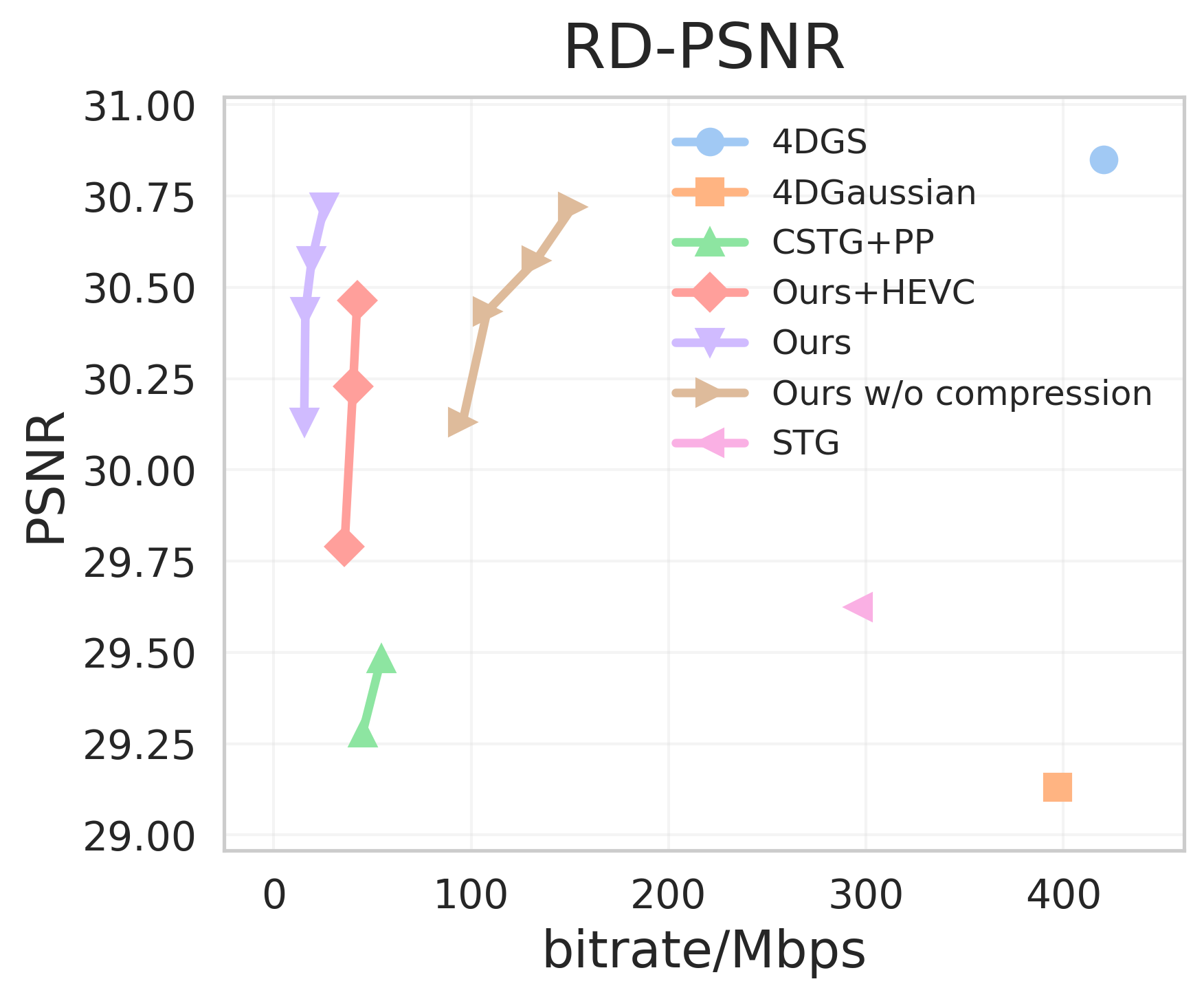}
    \end{subfigure}
    \begin{subfigure}{0.33\linewidth}
        \centering
        \parbox[][0.01cm][c]{1\linewidth}{\subsection*{~}\label{fig:RDSSIM2}}
        \includegraphics[width=1\linewidth]{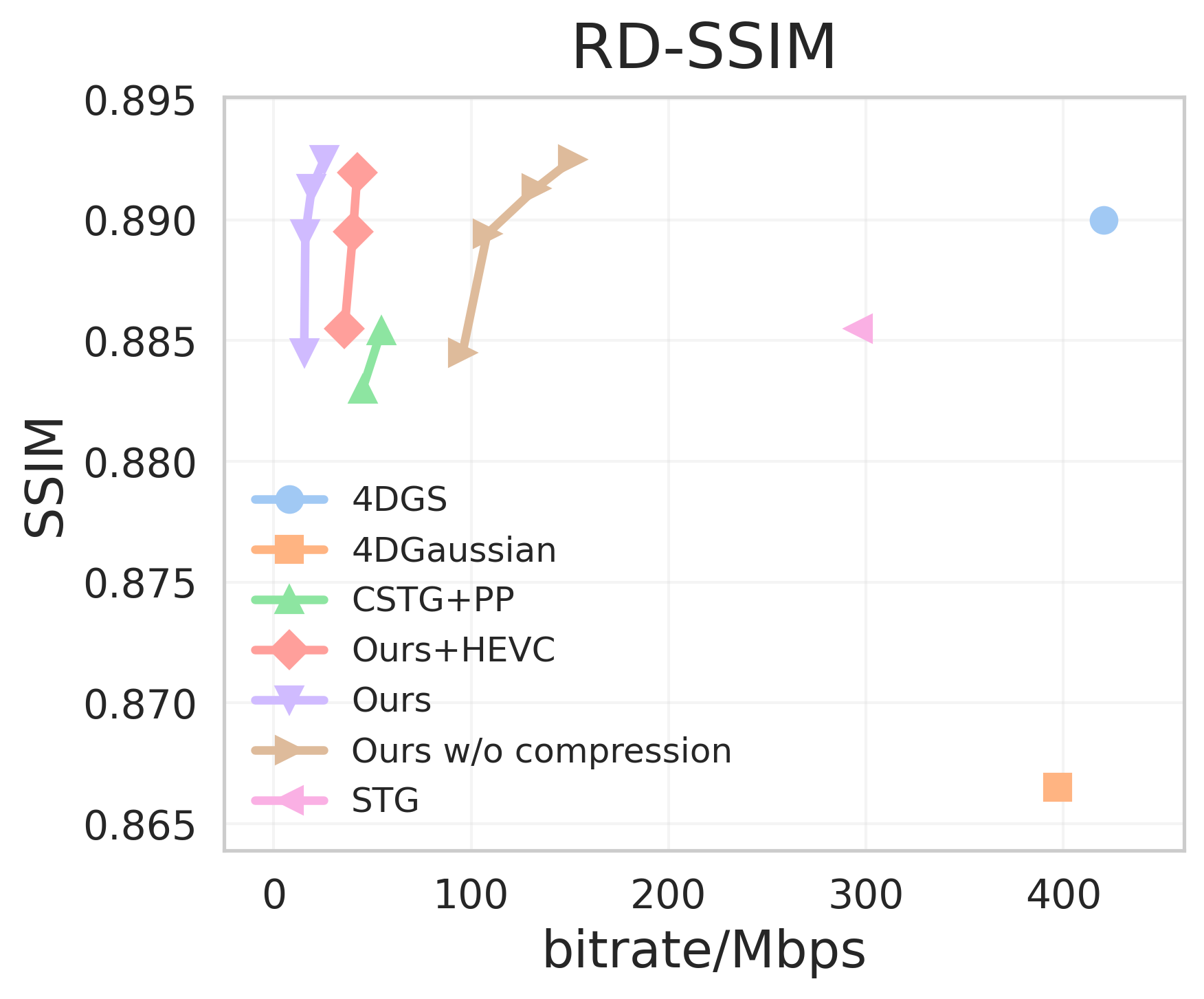}
    \end{subfigure}
    \begin{subfigure}{0.33\linewidth}
        \centering
        \parbox[][0.01cm][c]{1\linewidth}{\subsection*{~}\label{fig:RDLPIPS2}}
        \includegraphics[width=1\linewidth]{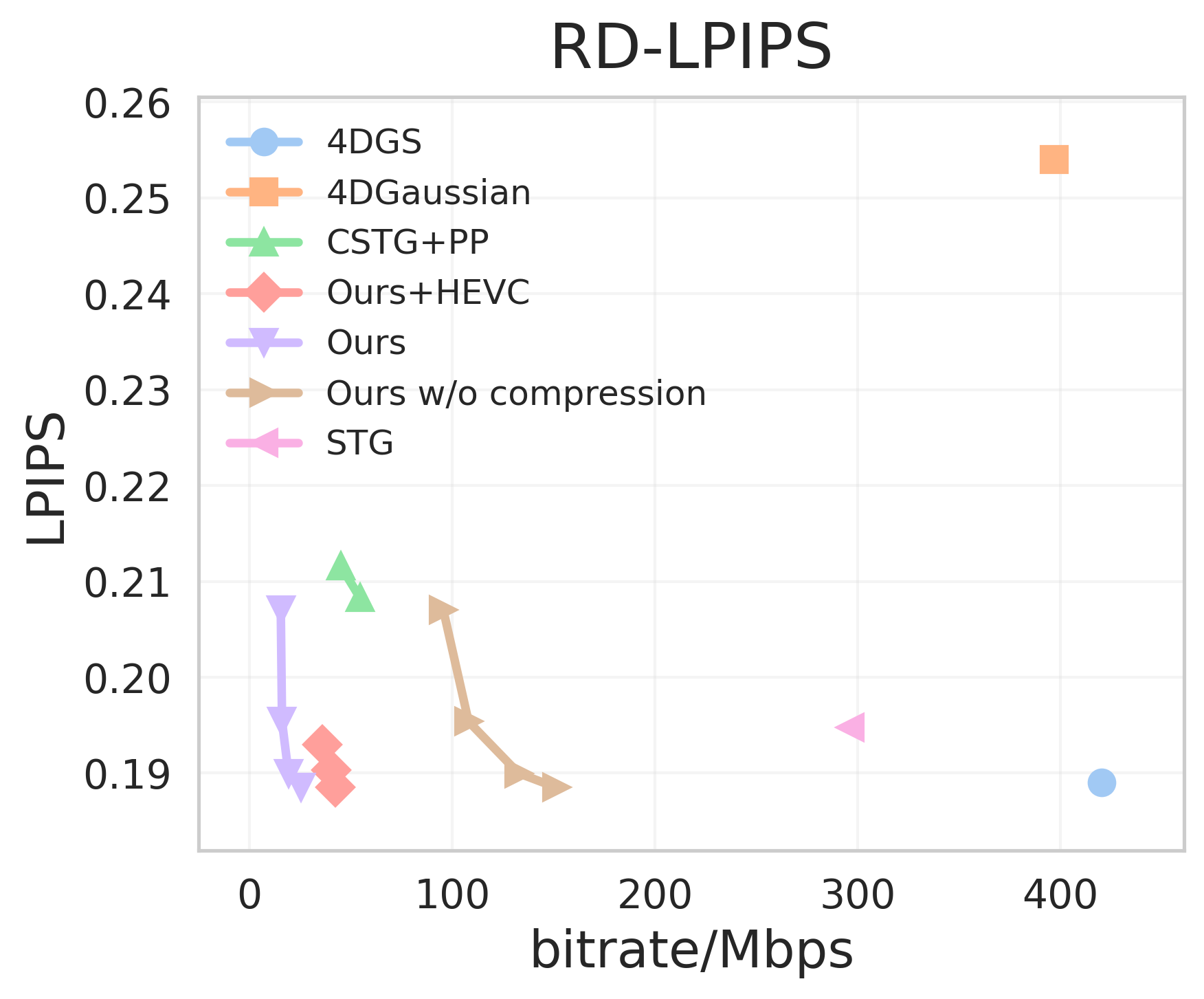}
    \end{subfigure}
    \caption{\textbf{RD Curve Comparision on MPEG dataset.} We visualize the RD Curve results in the \textbf{GOP 65} setting.}
    \label{fig:RD_Curve2}
\end{figure*}

We compress the features frames in an auto-regressive manner, grouping the frames into $n$ fragments, $\{\mathbf{F}_i\}_{i=1}^n$, and predicting their distribution as described in \cref{eq:dis_predict2}. This compression method effectively leverages both the channel and spatial dependencies of the anchors, as convolutional networks $h_{\phi}^\text{ent2}$ are employed to predict the conditional distribution.
\begin{equation}
\label{eq:dis_predict2}
    h_{\theta}^\text{ent2}: [\mathbf F_{i-k}; \cdots; \mathbf F_{i-1}] \mapsto \{\bmu_i, \boldsymbol{\sigma}_i\}
\end{equation}

We compress the $\bS_1,\bS_2,\{\bo^i\},M$ with the per-parameter distribution and adaptive quantization steps $\bQ$ predicted from time-independent feature $\bar\bff$ using neural network $h_{\phi}^\text{ent3}$:
\begin{align}
\label{eq:dis_predict3}
    &h_{\theta}^\text{ent3}: [\bar\bff] \mapsto \{\bmu, \boldsymbol{\sigma}, \bQ\}
\end{align}
The quantization of $\bS_1,\bS_2,\{\bo^i\},M$ is performed using the Straight-Through Estimator (STE):
\begin{equation}
    \bar{\boldsymbol{A}} = \text{SG}(\text{round}(\boldsymbol{A}/Q) - \boldsymbol{A}/Q)\times Q + \boldsymbol{A}
\end{equation}
Here, $\boldsymbol{A}$ represents the attributes in $\bS_1,\bS_2,\{\bo^i\},M$ and $\text{SG}$ denotes the stop gradient operation. The entropy loss for $\boldsymbol{A}$ is similar to that of $\bV_{GF}$, but the half quantization step is modified from 0.5 to $Q/2$.

For the position $\bx$ of anchors, we simply apply 16-bit quantization and store the results as a PNG image.

\section{Supplementary for Experiments}
\boldparagraph{Additional Implementation Details}
We implement 4DGaussian, 4DGS, STG and CSTG~\cite{wu20244d,yang2023real,li2024spacetime,lee2024compact2} using their official code bases. For the 4DGS, we only keep zero degrees of SH coefficients, therefore the storage will be much smaller than the original version. For the Neur3D dataset, we directly cite reported results from CSTG~\cite{lee2024compact2} in our paper. For the MPEG dataset, as no predefined hyper-parameters are available, we conduct multiple experiments with various hyper-parameter combinations, adjusted from the Neur3D configurations, and select the best-performing ones as our comparison benchmarks.

To obtain multi-rate results for CSTG on the MPEG dataset, we adjust the weight of the loss function associated with the pruning mask.

\boldparagraph{Per-Scene Quantitative Comparison}
Detailed per-scene results for the Neur3D dataset are provided in \tabref{tab:perscene_n3d}, and detailed per-scene results for the MPEG dataset are provided in \tabref{tab:perscene_mpeg}.

\begin{figure*}[t]
    \centering
    \includegraphics[width=\linewidth]{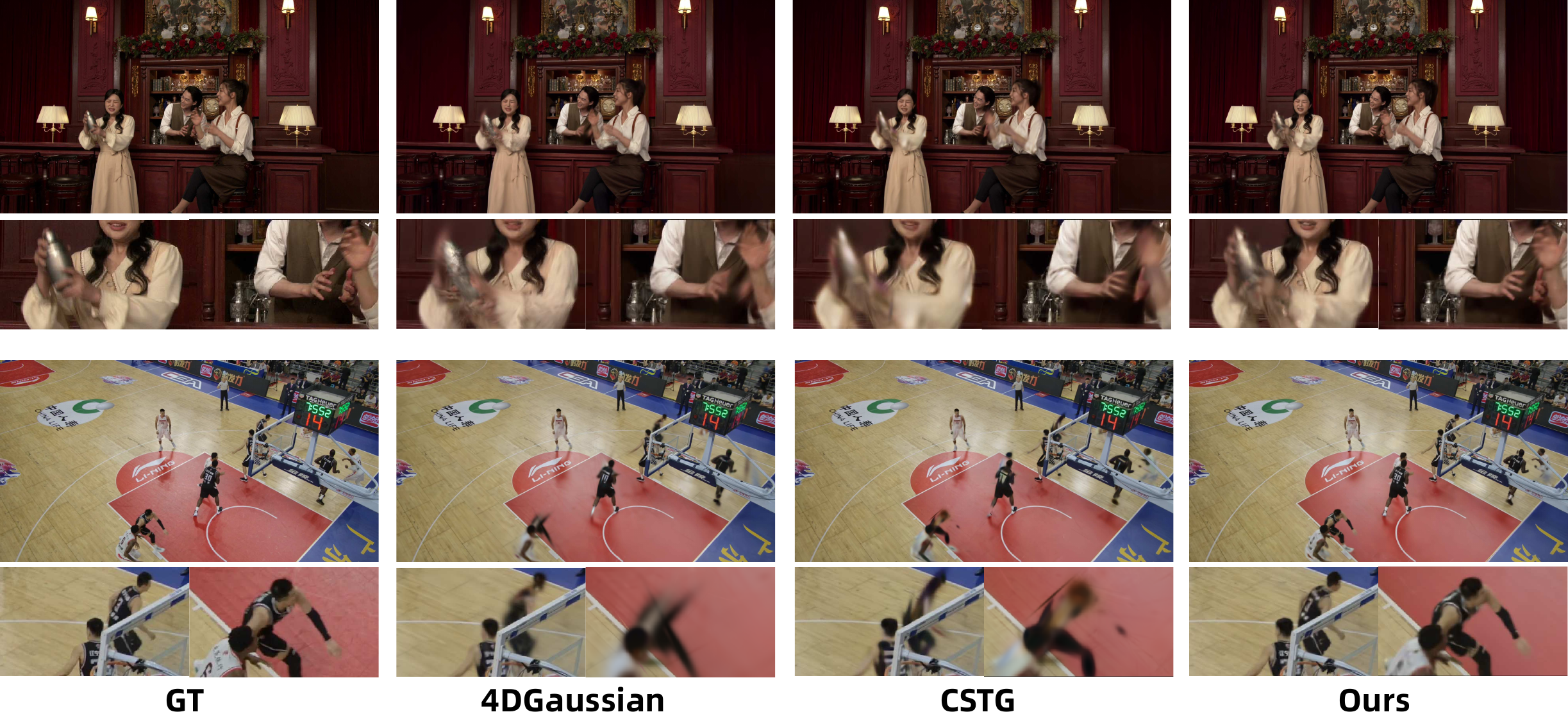}
    \caption{\textbf{More Qualitative Evaluation}. We present complete frames from the MPEG dataset along with a comparison of local details in this figure.}
    \label{fig:supp_vis2}
\end{figure*}

\begin{figure*}[t]
    \centering
    \includegraphics[width=\linewidth]{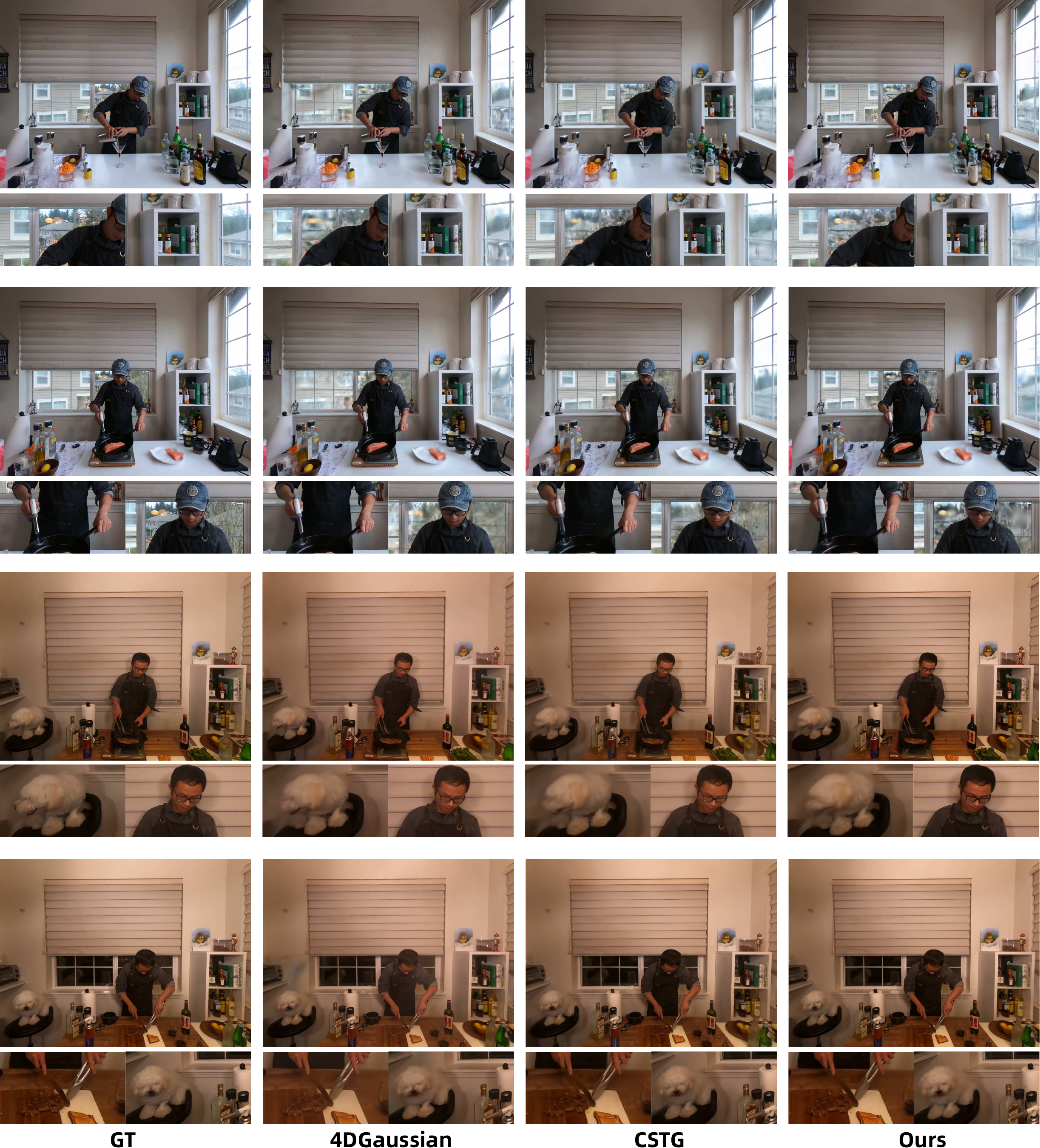}
    \caption{\textbf{More Qualitative Evaluation}. We present complete frames from the Neur3D dataset along with a comparison of local details in this figure.}
    \label{fig:supp_vis1}
\end{figure*}

\boldparagraph{More Qualitative Evaluation}
We provide additional qualitative results in \figref{fig:supp_vis1} and \figref{fig:supp_vis2} for better evaluation. To accurately segment the video into complete GOPs, We train GIFStream with a GOP of 60 for the Neur3D dataset and with a GOP of 65 for the MPEG dataset. Additionally, we observe flicker across different GOPs in the static background of the scenes like "flame salmon 1", since the reconstruction of content out of windows is challenging and artifacts in these areas look different in different GOPs. To improve the temporal consistency of these scenes, we utilize the previous GOP as initialization for the next GOP. Specifically, we load the checkpoint from the previous GOP and start training from the 10000th iteration for 10 minutes. During this process, we add and remove points to accommodate new content.

\begin{table}[t]
\centering
\resizebox{1.0\linewidth}{!}{
\begin{tabular}{lcccc}
\toprule
Scenes       & PSNR $\uparrow$ & SSIM $\uparrow$ & LPIPS $\downarrow$ & Storage (MB) $\downarrow$ \\\midrule

coffee martini      & 28.14 & 0.905 & 0.163 & 11.1   \\
cook spinach        & 33.03 & 0.950 & 0.138 & 12.0  \\
cut roasted beef    & 33.19 & 0.947 & 0.141 & 8.8   \\
flame salmon 1      & 28.51 & 0.916 & 0.157 & 8.2   \\
flame steak         & 33.76 & 0.957 & 0.134 & 8.2   \\
sear steak          & 33.83 & 0.958 & 0.134 & 10.2   \\
\bottomrule

\end{tabular}}
\caption{\textbf{Per-Scene Results on The Neur3D dataset}. We present the specific results of each scene on the Neur3D dataset.}
\label{tab:perscene_n3d}
\end{table}

\boldparagraph{Memory Break Down}
We provide the memory breakdown for parts of our experiments in \tabref{tab:breakdown}. This includes the memory usage of the compressed GIFStream using our end-to-end compression (GIFStream w/ Compression) and the uncompressed GIFStream, which is trained without compression techniques (GIFStream w/o Compression). Additionally, we present the increase in memory requirements when sparse feature masks are not applied (GIFStream w/o Sparse Mask $M_{de}$ ).

\boldparagraph{Compression Utilizing HEVC}
We perform simple compression utilizing image and video compression codecs on the MPEG dataset and present the RD-Curve in \figref{fig:RD_Curve2}. Specifically, we apply 16-bit quantization to the attributes $\mathbf{x},\bS_1,\bS_2,\{o^i\}_{i=1}^K,M$ and 8-bit quantization for the feature $\bff$ and $\bff_t$. Subsequently, we use PNG compression for $\mathbf{x},\bS_1,\bS_2,\{o^i\}_{i=1}^K,M,\bff$ and employ HEVC to compress $\bff_t$. While being inferior to our end-to-end compression solution, the HEVC-based compression also yields promising performance compared with other baselines. This demonstrates that our proposed feature streams are also compatible with existing video codecs.

\section{Limitation}
Our representation may exhibit inconsistencies in the background area between different GOPs, particularly in the distant background, where there are insufficient points for initialization. This is due to the instability of densification. The adaptive sampling strategy from STG~\cite{li2024spacetime} or the continual training approach can help alleviate this issue. Additionally, since our representation and compression methods require neural networks for inference, the computational demands may be unacceptable for some mobile devices.

\end{document}